\pdfoutput=1

\documentclass[11pt]{article}

\usepackage[final]{acl}
\usepackage{array}
\usepackage{multirow}
\usepackage{ragged2e}
\usepackage{booktabs}
\usepackage{arydshln}
\usepackage{times}
\usepackage{latexsym}
\usepackage{tabularx}
\usepackage[T1]{fontenc}
\usepackage{graphicx}
\usepackage{makecell}
\usepackage{enumitem}  
\usepackage[utf8]{inputenc}
\usepackage{enumitem}
\usepackage{microtype}

\usepackage{inconsolata}

\usepackage{graphicx}

\title{Enhancing Robustness of Retrieval-Augmented Language Models with In-Context Learning}

\author{
\textbf{Seong-Il Park},
\textbf{Seung-Woo Choi},
\textbf{Na-Hyun Kim},
\textbf{Jay-Yoon Lee} \\
Seoul National University\\
\texttt{\{athjk3,rhdn520,nahyun0410,lee.jayyoon\}@snu.ac.kr}}

\begin{document}
\maketitle
\begin{abstract}
Retrieval-Augmented Language Models (RALMs) have significantly improved performance in open-domain question answering (QA) by leveraging external knowledge. 
However, RALMs still struggle with unanswerable queries, where the retrieved contexts do not contain the correct answer, and with conflicting information, where different sources provide contradictory answers due to imperfect retrieval. 
This study introduces an in-context learning-based approach to enhance the reasoning capabilities of RALMs, making them more robust in imperfect retrieval scenarios. 
Our method incorporates Machine Reading Comprehension (MRC) demonstrations, referred to as \textit{cases}, to boost the model's capabilities to identify unanswerabilities and conflicts among the retrieved contexts. 
Experiments on two open-domain QA datasets show that our approach increases accuracy in identifying unanswerable and conflicting scenarios without requiring additional fine-tuning. 
This work demonstrates that in-context learning can effectively enhance the robustness of RALMs in open-domain QA tasks.
\end{abstract}
\section{Introduction}
\begin{figure}[ht]
    \centering
    \includegraphics[width=1\columnwidth]{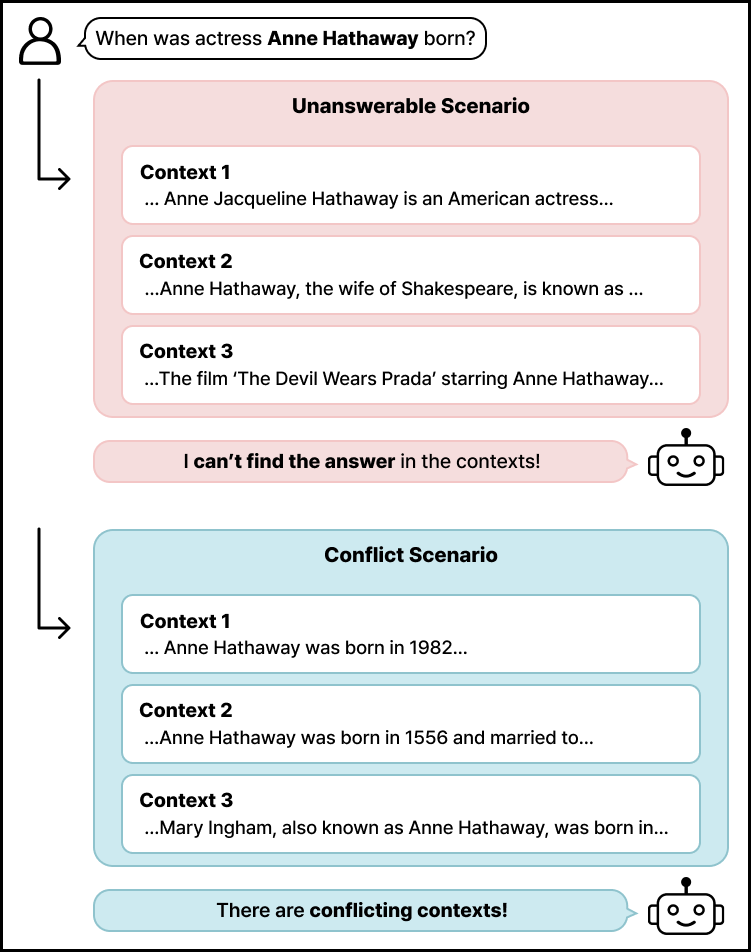}
    \caption{Examples of unanswerable and conflict scenario that may arise during retrieval-augmenation. A robust RALM should be able to identify such scenarios well.}
    \label{fig:intro}
\end{figure}
Retrieval Augmented Language Models (RALMs) have demonstrated remarkable performance in the field of open-domain question answering (QA). 
By leveraging external knowledge to generate answers, RALMs enhance accuracy and enable language models to respond to queries beyond their training data. \citep{lewis2020retrieval,guu2020retrieval,izacard2021leveraging,izacard2022few} 
Typically, RALMs operate in two stages: the retrieval step, which involves fetching relevant contexts from external knowledge sources, and the generation step, where answers are generated based on the retrieved contexts. 
Recent research has shown that using frozen Large Language Models (LLMs) without additional fine-tuning during the generation step can also be effective. \citep{ram2023context, shi2023replug}

However, a critical issue in open-domain QA is the reliance of RALMs on the quality of external knowledge. 
Figure \ref{fig:intro} illustrates common imperfect retrieval scenarios in RALMs. 
In unanswerable scenario where the retrieved contexts do not contain the correct answer, RALMs cannot provide an accurate response. 
Additionally, when contexts are retrieved from various sources, such as search engines, conflicting information may arise. 
In such scenario, RALMs may struggle to determine the correct information, leading to reliance on their parametric knowledge or potential hallucination.
\begin{figure*}[]
    \centering
    \includegraphics[width=1.0\textwidth]{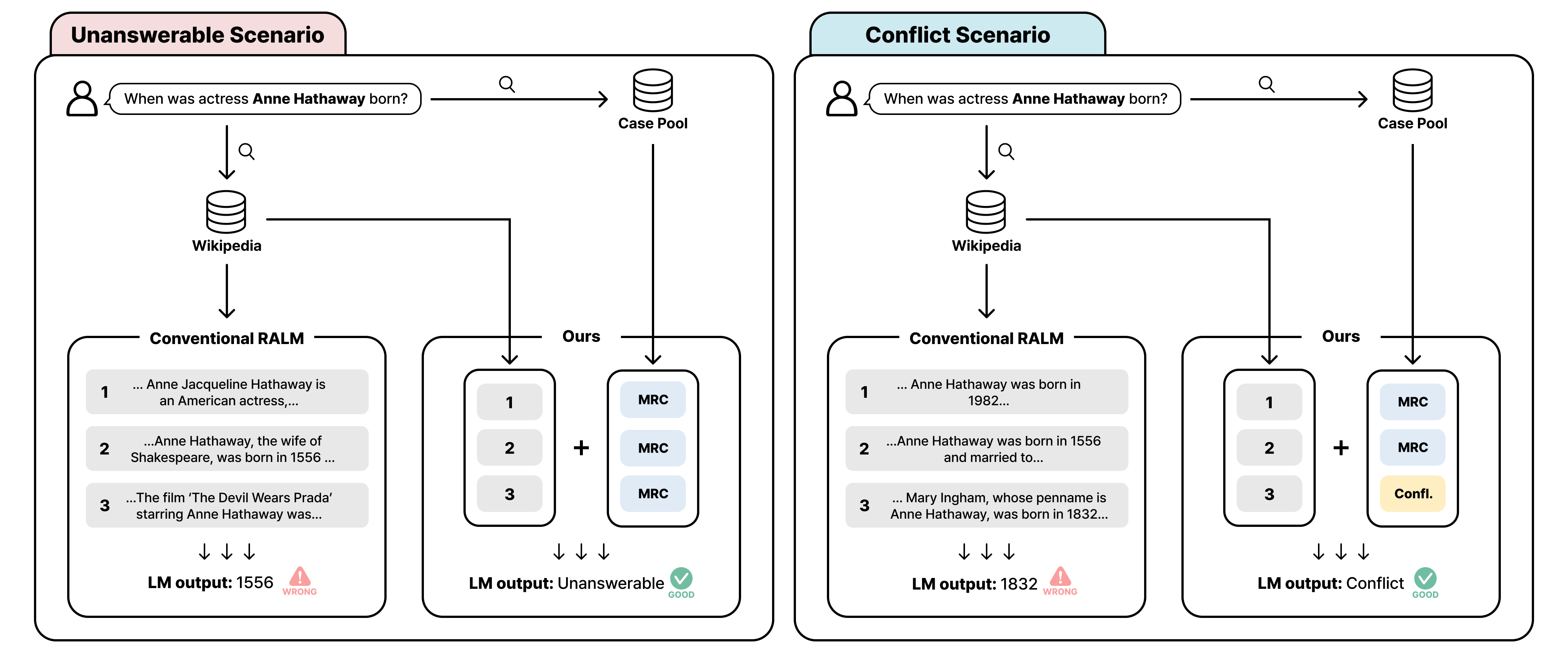}
    \caption{An overview of our approach. Conventional RALM generates answers by providing the LLM with context retrieved from a knowledge source. In contrast, our method simultaneously inputs cases that enhance the LLM's reasoning capability, allowing it to generate answers. This leads to more robust reasoning compared to conventional RALM.}
    \label{fig:main-fig}
\end{figure*}

To address these challenges, we propose the in-context learning \citep{brown2020language} based approach to enhance the reasoning capabilities of LLMs, thereby increasing robustness in such imperfect retrieval scenarios. 
Unlike previous approaches that depend on extensive fine-tuning \citep{chen2022rich, asai2023self, yoran2023making, yu2023chain, neeman2023disentqa}, our method leverages the in-context learning capability of LLMs, demonstrating that providing simple examples to LLMs can improve robustness in open-domain QA without additional training. 
Figure \ref{fig:main-fig} provides an overview of our approach. 
Unlike conventional RALM, our method retrieves demonstrations (referred to as cases) that assist in answering a given query. 
By concatenating these retrieved cases to the LLM's input during retrieval-augmentation, we enhance the LLM's reasoning abilities through in-context learning. 
This enables the RALMs to perform more robust reasoning.

Our experiments show that providing LLMs with Machine Reading Comprehension (MRC) demonstrations enhances accuracy and the ability to detect unanswerability. 
Additionally, presenting LLMs with simple examples that simulate conflicts among retrieved contexts improves their ability to identify such conflicts.

Our contributions and key findings are summarized as follows:
\begin{itemize}[noitemsep]
    \item We demonstrated that providing RALMs with MRC demonstrations improves their reasoning capabilities in open-domain QA, where answers should be generated from multiple documents.
    \item Using retrieval to select similar demonstrations is more effective than randomly selecting those from the entire pool.
    \item Providing QA cases alone enhances reasoning and improves robustness in scenarios with frequently encountered issues in open-domain QA, such as unanswerable queries.
    \item For conflict scenario that LLMs do not frequently encounter during training, directly providing analogous demonstrations improves reasoning abilities.
\end{itemize}

\section{Related Works}
\subsection{In-context learning and RALMs}
Large Language Models (LLMs) have demonstrated the ability to learn from a few examples in their immediate context, a capability known as in-context learning (ICL). 
This capability, widely recognized as an emerging trait in many advanced models, focuses on gaining knowledge through inference \citep{brown2020language, wei2022chain}. 
In open-domain QA, recent works highlighted that appending relevant documents to LLMs’ inputs without additional training significantly enhanced performance, providing an efficient method for RALMs \citep{ram2023context}. 
Similarly, \citep{shi2023replug} applied retrieval-augmented methods to black-box language models, enhancing their question-answering capabilities without altering their internal structure. 
Another study introduces Fusion-in-Context, which examined how various prompting strategies influence few-shot learning performance \citep{huang2023raven}. 
Following these approaches, we enhance the RALMs' robustness using in-context learning methods.
\subsection{Robustness of RALMs on unanswerability}
Various studies have aimed to increase the robustness of RALMs in unanswerable scenarios. \citep{yu2023chain} introduced the Chain-Of-Note, which trains LLMs to generate answers after assessing the relevance of retrieved documents through sequential reading notes. \citep{yoran2023making} trained RALMs to handle unanswerability using an automatically generated dataset. Self-RAG \citep{asai2023self} generated special tokens to indicate the relevance of retrieved documents or the need for further retrieval. CRAG \citep{yan2024corrective} used a lightweight retrieval evaluator to assess unanswerability. While these approaches have improved robustness, leveraging LLMs' in-context learning capabilities in these scenarios is still underexplored.
\subsection{Robustness of RALMs on conflicts}
Knowledge conflicts can arise from clashes between parametric and contextual knowledge \citep{longpre2021entity} or among various contextual knowledges \citep{chen2022rich}. 
Previous studies have focused on training models to prioritize contextual knowledge, disentangle knowledge types \citep{neeman2023disentqa} or measure decision-making patterns \citep{ying2023intuitive}. 
Several studies have also aimed to mitigate conflicts by calibrating models to answer only when there's no conflict \citep{chen2022rich}, searching for diverse passages by augmenting queries \citep{weller2022defending}, or filtering out conflicting passages \citep{hongso}. 
However, these approaches often overlook the LLMs' in-context learning capabilities. Unlike previous works, we focus on leveraging the model's in-context learning to make it \textit{conflict-awarable} for more reliable outputs without additional training.

\section{Method}
Our objective is to enhance the reasoning capabilities of LLMs in open-domain QA scenarios, particularly in detecting unanswerable scenarios where no answer exists within the retrieved contexts, and conflict scenarios where contradictions exist among retrieved contexts.

Our approach follows the In-context RALM method \citep{ram2023context}, which concatenates retrieved contexts as inputs to a frozen LLM for retrieval-augmentation.
To further enhance the LLM’s reasoning capability, we will add demonstrations to the RALMs by simply concatenating demonstrations to the existing RALM input. 
Typically, in-context learning provides examples of the same task \citep{dong2022survey}, but our demonstrations are based on Machine Reading Comprehension (MRC) datasets, which have a single shorter context, rather than generating answers from multiple documents as in ODQA. 
We refer to these demonstrations as \textbf{\textit{cases}}.
\subsection{Crafting cases}
\label{subsec:crating_case}
We create a case pool using the SQuAD \citep{rajpurkar2016squad}, which is a well-known MRC dataset consisting of question, context, and answer pairs. 
From this dataset, we create two types of cases:

\textbf{QA case} To improve reasoning capability and unanswerability detection in open-domain QA, we use MRC examples as QA cases. 
Given that open-domain QA resembles an MRC task involving multiple documents, we use SQuAD examples without additional perturbation, excluding those with lengthy contexts \footnote{We filtered out contexts containing more than 150 words.}.

\textbf{Conflict case} We follow the method by \citep{xie2023adaptive} to create conflict cases. While \citet{xie2023adaptive} created counter memories contradicting the LLM’s parametric knowledge, we create conflicting contexts contradicting the retrieved contexts. The process is as follows:
\begin{enumerate}[noitemsep,partopsep=0pt]
\item \textbf{Answer Sentence Creation:} Similar to \citet{xie2023adaptive}, we generate base sentences for entity substitution using the question and answers from open-domain QA datasets, forming declarative answer sentences. 
We utilize an LLM for this step.
\item \textbf{Entity Substitution and Filtering:} We substitute the answer entity in the answer sentence with another entity of the same type, creating a conflict sentence. 
Then, using an LLM, we generate a conflict passage supporting the conflict sentence. 
Any conflict passage containing the answer string is excluded.
\item \textbf{Concatenation:} By concatenating the conflict passage with the original context, we simulate a scenario with multiple contradicting documents, creating a conflict case.
\end{enumerate}
We use the \texttt{Llama3-70B-Instruct} \citep{touvron2023llama} for generating cases. For entity substitution, we use SpaCy NER model for entity recognition.\footnote{We used the \texttt{en\_core\_web\_trf} model. The entities for substitutions were created by extracting entities from all texts in the \texttt{Wikitext-103-raw-v1}.} Details on prompts and settings used for the LLM are provided in Appendix \ref{sec:prompt}.
\subsection{Case retrieval}
\label{subsec:case_retrieval}
At inference time, we put the crafted cases into the LLM. 
Similar to \citep{thai2023machine}, we employ a case-based reasoning method for case selection. 
We mask entities in the test set questions (referred to as queries) and case set questions, compute sentence embeddings\footnote{For sentence embedding, we used \texttt{all-MiniLM-L6-v2} model from Sentence Transformers library \citep{reimers2019sentence}} for the masked questions, and calculate cosine similarity between these embeddings. 
The top-k similar cases are used as demonstrations during inference, enabling effective in-context learning by providing the LLM with cases similar to the current query. 
To prevent leakage due to cases, any case where the answer matched the query answer is excluded from the case candidates.

\section{Experimental Setup}
\subsection{Dataset}
We used the Natural Questions (NQ) \citep{lee2019latent} and Web Questions (WebQ) \citep{berant2013semantic} datasets, commonly employed in open-domain QA tasks. 
Both datasets' test sets were used for our experiments. 
We retrieved the top five documents for each query from Wikipedia\footnote{We used the preprocessed data from \citep{karpukhin2020dense}} based on their cosine similarity.
For dense retriever, we use ColBERTv2 \citep{santhanam2022colbertv2} to retrieve most similar contexts for each query. 
Detailed statistics for each dataset are provided below.

To simulate unanswerable and conflict scenarios, we perturbed the existing open-domain QA datasets to create unanswerable and conflict test sets.
\begin{table*}[ht]
\centering\scriptsize
\setlength{\tabcolsep}{10pt}
\renewcommand{\arraystretch}{1}
\resizebox{\textwidth}{!}{\begin{tabular}{clccc|ccc}\Xhline{2\arrayrulewidth}
 & \multicolumn{1}{c}{} & \multicolumn{3}{c|}{\textbf{NQ}} & \multicolumn{3}{c}{\textbf{WebQ}} \\
Model & \multicolumn{1}{c}{Prompt} & Acc & \begin{tabular}[c]{@{}c@{}}Acc\\ (ans)\end{tabular} & \begin{tabular}[c]{@{}c@{}}Acc\\ (unans)\end{tabular} & Acc & \begin{tabular}[c]{@{}c@{}}Acc\\ (ans)\end{tabular} & \begin{tabular}[c]{@{}c@{}}Acc\\ (unans)\end{tabular} \\ \hline
Llama3 & \multicolumn{1}{l|}{zeroshot} & 52.97 & 58.83 & 35.61 & 35.00 & 39.54 & 22.30 \\
 & \multicolumn{1}{l|}{1Q} & 54.12 & 60.01 & 36.65 & 36.33 & 41.72 & 21.28 \\
 & \multicolumn{1}{l|}{3Q} & 56.84 & 62.67 & 39.54 & 39.80 & 45.59 & 23.65 \\
 & \multicolumn{1}{l|}{5Q} & \textbf{57.15} & \textbf{62.67} & \textbf{40.79} & \textbf{43.99} & \textbf{49.70} & \textbf{28.04} \\ \hline
Qwen1.5 & \multicolumn{1}{l|}{zeroshot} & 58.19 & 67.34 & 31.06 & 48.80 & 58.52 & 21.62 \\
 & \multicolumn{1}{l|}{1Q} & 59.34 & 65.95 & 39.75 & 48.80 & 58.16 & 22.64 \\
 & \multicolumn{1}{l|}{3Q} & \textbf{60.96} & 67.34 & \textbf{42.03} & \textbf{50.85} & 59.01 & \textbf{28.04} \\
 & \multicolumn{1}{l|}{5Q} & 60.23 & \textbf{67.90} & 37.47 & 50.31 & \textbf{60.10} & 22.97 \\ \hline
ChatGPT & \multicolumn{1}{l|}{zeroshot} & 42.48 & 41.45 & 45.55 & 27.96 & 26.72 & 31.42 \\
 & \multicolumn{1}{l|}{1Q} & 47.03 & 41.52 & 63.35 & 33.75 & 26.96 & 52.70 \\
\multicolumn{1}{l}{} & \multicolumn{1}{l|}{3Q} & \textbf{48.80} & 42.57 & \textbf{67.29} & 34.28 & 26.12 & \textbf{57.09} \\
\multicolumn{1}{l}{} & \multicolumn{1}{l|}{5Q} & 47.96 & \textbf{43.06} & 62.53 & \textbf{34.55} & \textbf{28.42} & 51.69 \\ \hline
\Xhline{1\arrayrulewidth}
\end{tabular}}
\caption{Experimental results on unanswerable set. In the prompt column, "XQ" denotes that X QA cases have been added. Acc represents the accuracy on all examples, Acc (ans) indicates the accuracy on answerable examples, and Acc (unans) shows the accuracy on unanswerable examples. The best performance is highlighted in bold.}
\label{tab:unans}
\end{table*}

\textbf{Unanswerable Set} To determine if a query is answerable based on retrieved contexts, we use both string match and an NLI model\footnote{We used \texttt{MoritzLaurer/mDeBERTa-v3-base-xnli-mul-\ \ tilingual-nli-2mil7} from Hugging Face transformers library}. If the retrieved context does not contain the answer string and the context-query pair is not entailed, we consider the context unanswerable. 
If all top-k retrieved contexts are unanswerable, the query is labeled as an unanswerable example and the original answer is replaced with \textit{unanswerable}.

\textbf{Conflict Set} We utilized the method described in the \ref{subsec:crating_case} to create a conflict passage for each query, which is then randomly inserted among the top five retrieved contexts to generate conflict examples. 
To differentiate between the cases and the test set, we employed the \texttt{GPT-3.5-turbo-0125} model for generating conflict passages. 
To occur a conflict, the original top five retrieved contexts must contain the correct answer, hence we inserted the conflict passages only into answerable examples. 
To determine answerability, similar to the unanswerable set, we considered a context as answerable if it included the answer string and the question-context pair was entailment. 
If at least one answerable context existed among the top-k retrieved contexts, the example was considered answerable.
After inserting a single conflict passage into the answerable example, the original answer is replaced with the label \textit{conflict}, similar to the process used for the unanswerable set.

These perturbations allow us to evaluate the effectiveness of our method in improving the LLM’s ability to handle unanswerable and conflicting scenarios in open-domain QA.

\subsection{Prompting}
We designed instructions to evaluate how well RALMs can identify unanswerability and conflicts in the unanswerable and conflict sets, respectively.
These instructions are designed to extend standard retrieval-augmented QA by adding the capability to identify unanswerable and conflicting contexts. 
Prompts for each type are as follows:

\textbf{Unanswerable Prompt} This instruction adds the task of identifying unanswerability. 
The LLM must provide an answer for answerable examples and respond with \textit{unanswerable} if the context does not contain the answer.

\textbf{Conflict Prompt} This instruction adds the task of identifying conflicts among contexts. 
The LLM have to respond with \textit{conflict} if there is contradiction among the retrieved contexts and provide an answer if there is no contradiction.

Please refer to the Appendix \ref{sec:prompt} for the details of the prompt.

\subsection{Metric}
Following \citep{mallen2023not}, we used accuracy as our metric. Unlike exact match, accuracy considers a response correct if it contains the answer string. 
To prevent distortion due to long responses, we limited the response length to 10 tokens during generation.
\subsection{LLM}
For effective in-context learning, we used models with large parameter sizes. 
Specifically, we used the Llama3-70B-Instruct model \citep{touvron2023llama}, the Qwen-1.5-chat-72B model \citep{bai2023qwen} and the \texttt{GPT-3.5-turbo-0125} model (abbreviated as ChatGPT) using OpenAI's API. 
To reduce generation randomness, we used greedy decoding and fixed the random seed.
For faster inference, we used vLLM \citep{kwon2023efficient}.
\section{Experiments}
In these experiments, we aim to investigate how effectively our constructed cases can help LLMs identify unanswerability and conflicts in open-domain QA scenarios.
\begin{table*}[ht]
\centering
\centering\scriptsize
\setlength{\tabcolsep}{10pt}
\renewcommand{\arraystretch}{1}
\resizebox{\textwidth}{!}{\begin{tabular}{llccc|ccc}\Xhline{2\arrayrulewidth}
 &  & \multicolumn{3}{c|}{\textbf{NQ}} & \multicolumn{3}{c}{\textbf{WebQ}} \\
\multicolumn{1}{c}{Model} & \multicolumn{1}{c}{Prompt} & \begin{tabular}[c]{@{}c@{}}Acc\\ (NC)\end{tabular} & \begin{tabular}[c]{@{}c@{}}Acc\\ (C)\end{tabular} & \begin{tabular}[c]{@{}c@{}}Acc\\ (Avg)\end{tabular} & \begin{tabular}[c]{@{}c@{}}Acc\\ (NC)\end{tabular} & \begin{tabular}[c]{@{}c@{}}Acc \\ (C)\end{tabular} & \begin{tabular}[c]{@{}c@{}}Acc\\ (Avg)\end{tabular} \\ \hline
Llama3 & \multicolumn{1}{l|}{zeroshot} & 58.54 & 10.67 & 34.61 & 38.55 & 8.75 & 23.65 \\
 & \multicolumn{1}{l|}{1Q} & 64.61 & 16.18 & 40.39 & 42.27 & 14.53 & 28.40 \\ \cline{2-8} 
 & \multicolumn{1}{l|}{3Q} & 70.79 & 15.28 & 43.03 & 42.83 & 8.01 & 25.42 \\
 & \multicolumn{1}{l|}{2Q+1C} & \textbf{71.24} & \textbf{25.73} & \textbf{48.48} & \textbf{50.65} & \textbf{28.12} & \textbf{39.39} \\ \cline{2-8} 
 & \multicolumn{1}{l|}{5Q} & \underline{\textbf{72.81}} & 24.38 & 48.60 & 50.65 & 13.04 & 31.84 \\
 & \multicolumn{1}{l|}{3Q+2C} & 71.01 & \underline{\textbf{35.17}} & \underline{\textbf{53.09}} & \underline{\textbf{51.77}} & \underline{\textbf{35.38}} & \underline{\textbf{43.58}} \\ \hline
Qwen1.5 & \multicolumn{1}{l|}{zeroshot} & \underline{76.29} & 8.76 & 42.53 & 59.59 & 13.04 & 36.31 \\
 & \multicolumn{1}{l|}{1Q} & 71.35 & 12.70 & 42.02 & 58.10 & 21.79 & 39.94 \\ \cline{2-8} 
 & \multicolumn{1}{l|}{3Q} & 73.26 & 19.78 & 46.52 & \underline{\textbf{59.96}} & 22.91 & 41.43 \\
 & \multicolumn{1}{l|}{2Q+1C} & 73.03 & \underline{\textbf{25.28}} & \underline{\textbf{49.16}} & 57.54 & \underline{\textbf{32.77}} & \underline{\textbf{45.16}} \\ \cline{2-8} 
 & \multicolumn{1}{l|}{5Q} & \textbf{74.04} & 16.63 & 45.34 & \textbf{58.66} & 24.95 & 41.81 \\
 & \multicolumn{1}{l|}{3Q+2C} & 73.60 & \textbf{24.16} & \textbf{48.88} & 57.73 & \textbf{27.93} & \textbf{42.83} \\ \hline
ChatGPT & \multicolumn{1}{l|}{zeroshot} & 55.51 & 28.65 & 42.08 & 34.82 & 38.36 & 36.59 \\
 & \multicolumn{1}{l|}{1Q} & 52.81 & 28.76 & 40.79 & 34.64 & 40.60 & 37.62 \\ \cline{2-8} 
 & \multicolumn{1}{l|}{3Q} & \textbf{58.20} & 29.21 & \textbf{43.71} & 37.80 & \underline{\textbf{42.46}} & \underline{\textbf{40.13}} \\
 & \multicolumn{1}{l|}{2Q+1C} & 57.08 & \textbf{29.89} & 43.48 & 37.62 & 40.41 & 39.01 \\ \cline{2-8} 
 & \multicolumn{1}{l|}{5Q} & \underline{\textbf{58.65}} & 23.71 & 41.18 & \underline{\textbf{41.71}} & 38.18 & \textbf{39.94} \\
 & \multicolumn{1}{l|}{3Q+2C} & 56.85 & \underline{\textbf{31.57}} & \underline{\textbf{44.21}} & 38.18 & \textbf{40.78} & 39.48 \\ \hline
\Xhline{1\arrayrulewidth}
\end{tabular}}
\caption{Experimental results on conflict set. In the prompt column, + indicates that two case were used together. Acc (NC) denotes the accuracy on non-conflict examples, Acc (C) represents the accuracy on conflict examples, and Acc (Avg) is the average accuracy of the two. The best performance for each total case count is highlighted in bold, and the overall best performance is underlined.}
\label{tab:conflict}
\end{table*}

\subsection{Experiments on Unanswerable Set}
Table \ref{tab:unans} presents the results of our experiments on identifying unanswerable questions based on different types of prompts. 
The number preceding the case name indicates the number of added cases.
Our goal is not only to have LLMs correctly identify unanswerable examples but also to ensure them to provide accurate answers for answerable examples. 
Therefore, we calculated the accuracy for both unanswerable and answerable examples, as well as the overall accuracy. 
These results indicate that adding QA cases consistently enhance the reasoning capabilities of LLMs across all models and datasets. 
Specifically, the accuracy for unanswerable examples significantly increased compared to the zero-shot performance. 
For instance, ChatGPT showed an improvement of up to 21.74 in the NQ dataset and 25.67 in the Web Questions dataset. 
This improvement indicates that providing QA cases enhances the LLMs' ability to reason in situations where no correct answer exists. 
However, the impact of adding QA cases varied among models. 
For example, Llama3's performance continued to improve with more QA cases, while Qwen1.5 achieved the best performance with three QA cases. 
These findings imply that simple examples can significantly boost the reasoning abilities of LLMs through in-context learning.

\subsection{Experiments on Conflict Set}
Unlike the unanswerable experiments, we include both QA and conflict cases in our conflict set experiments, while keeping the total number of cases constant for fair comparison. 
Table \ref{tab:conflict} shows the results of our experiments on identifying conflicts. When using both QA and conflict cases, we first added the QA cases, followed by the conflict cases in the prompt.
To evaluate the LLMs' ability to identify conflicts while maintaining accuracy on answerable examples, we conducted two forward passes. 
In the first pass, we inferred the answerable examples without adding conflict passages (non-conflict examples, abbreviated as \textbf{NC}). 
In the second pass, we add conflict passages to the same examples (conflict examples, abbreviated as \textbf{C}) and then inferred. 
We calculated the accuracy for both passes to assess the models' performance in identifying conflicts and answering correctly.
The results show that adding QA cases alone improves accuracy on conflict examples compared to zero-shot performance. 
Moreover, adding appropriate conflict cases provides even more benefits. 
Model performance varied; for example, Qwen showed the highest accuracy for non-conflict examples in the zero-shot setting but had lower accuracy for conflict examples, with the best overall performance achieved using a combination of \textbf{2Q+1C}. 
Conversely, Llama3 performed best with the \textbf{3Q+2C} combination, except for the \textbf{5Q} setting. 
ChatGPT's conflict accuracy improved with added conflict cases, but its accuracy for non-conflict examples decreased compared to adding only QA cases. 
Additionally, ChatGPT showed less improvement in conflict example accuracy compared to other models when conflict cases were added. 
These results are discussed in more detail in \ref{conflict_study}.

Overall, the experiments indicate that identifying conflicts requires more complex reasoning than identifying unanswerable, and the effect of adding QA cases alone is limited. 
However, providing simplified examples that mimic more complex scenarios can enhance reasoning capabilities. 
This suggests that simple examples can significantly improve the robustness of LLMs without additional fine-tuning. Also, it shows that such direct examples, like conflicts which are difficult for LLMs to encounter during training, can be more effective in improving reasoning abilities.
\begin{table}[hb]
\resizebox{\columnwidth}{!}{%
\begin{tabular}{llcccc}
\hline
\multicolumn{1}{c}{\textbf{Model}} & \multicolumn{1}{c}{\textbf{Method}} & \textbf{Size} & \textbf{Acc} & \textbf{\begin{tabular}[c]{@{}c@{}}Acc\\ (ans)\end{tabular}} & \textbf{\begin{tabular}[c]{@{}c@{}}Acc\\ (unans)\end{tabular}} \\ \hline
\multicolumn{1}{l|}{\multirow{6}{*}{ChatGPT}} & \multicolumn{1}{l|}{Ours} & 1 & 47.03 & 41.52 & 63.35 \\
\multicolumn{1}{l|}{} & \multicolumn{1}{l|}{Random} & 1 & 44.10 & 36.92 & 65.42 \\ \cline{2-6} 
\multicolumn{1}{l|}{} & \multicolumn{1}{l|}{Ours} & 3 & 48.80 & 42.57 & 67.29 \\
\multicolumn{1}{l|}{} & \multicolumn{1}{l|}{Random} & 3 & 44.94 & 36.50 & 69.98 \\ \cline{2-6} 
\multicolumn{1}{l|}{} & \multicolumn{1}{l|}{Ours} & 5 & 47.96 & 43.06 & 62.53 \\
\multicolumn{1}{l|}{} & \multicolumn{1}{l|}{Random} & 5 & 43.74 & 37.82 & 61.28 \\ \hline
\multicolumn{1}{l|}{\multirow{6}{*}{Llama3}} & \multicolumn{1}{l|}{Ours} & 1 & 54.12 & 60.01 & 36.65 \\
\multicolumn{1}{l|}{} & \multicolumn{1}{l|}{Random} & 1 & 53.13 & 58.76 & 36.44 \\ \cline{2-6} 
\multicolumn{1}{l|}{} & \multicolumn{1}{l|}{Ours} & 3 & 56.84 & 62.67 & 39.54 \\
\multicolumn{1}{l|}{} & \multicolumn{1}{l|}{Random} & 3 & 54.23 & 59.32 & 39.13 \\ \cline{2-6} 
\multicolumn{1}{l|}{} & \multicolumn{1}{l|}{Ours} & 5 & 59.13 & 64.20 & 44.10 \\
\multicolumn{1}{l|}{} & \multicolumn{1}{l|}{Random} & 5 & 57.93 & 62.25 & 45.13 \\ \hline
\end{tabular}%
}

\caption{Experimental results on the unanswerable set of NQ. Method refers to the case retrieval approach, and size denotes the number of added cases. Acc represents the accuracy on all examples, Acc (ans) indicates the accuracy on answerable examples, and Acc (unans) represents the accuracy on unanswerable examples.}
\label{tab:retrieval}
\end{table}

\subsection{Further Analysis}
\subsubsection{Case Selection}
To verify the effectiveness of our case retrieval method described in \ref{subsec:case_retrieval}, we compared the results of selecting cases using our method versus randomly selecting cases from the entire pool. 
Table \ref{tab:retrieval} shows the results for the NQ unanswerable set. Our method demonstrates higher overall accuracy compared to randomly selecting cases. 
Specifically, for answerable examples, our method achieves up to 6 higher accuracy. 
This indicates that our case retrieval approach may be an effective strategy for in-context learning.
\subsubsection{Impact of Conflict Cases on ChatGPT}
\label{conflict_study}
We conducted additional experiments to understand why adding conflict cases to ChatGPT is less effective. We calculated the False Conflict Detection Rate (FCDR), which is the rate at which non-conflict examples are incorrectly predicted as "conflict," for each model. We compared the results of zeroshot and with three additional QA cases. The results are shown in Table \ref{tab:false_detection}. ChatGPT exhibits a significantly higher FCDR compared to Llama3 and Qwen1.5, with 17.08 on NQ and 25.33 on WebQ in the zeroshot setting. This rate further increases to 20.79 and 30.17, respectively, when additional QA cases are included. This suggests that ChatGPT has been trained to be more sensitive to conflicts, which limits the improvement in accuracy for conflict examples when more conflict cases are added. These findings indicate that the effectiveness of case additions can vary depending on the model's characteristics, which we will leave for future work.
\begin{table}[]
\centering\tiny
\resizebox{\columnwidth}{!}{%
\begin{tabular}{clcc}
\hline
\textbf{Model} & \multicolumn{1}{c}{\textbf{Prompt}} & \textbf{NQ} & \textbf{WebQ} \\ \hline
\multicolumn{1}{c|}{ChatGPT} & \multicolumn{1}{l|}{zeroshot} & 17.08 & 25.33 \\
\multicolumn{1}{l|}{\textbf{}} & \multicolumn{1}{l|}{QA} & 20.79 & 30.17 \\ \hline
\multicolumn{1}{c|}{Llama3} & \multicolumn{1}{l|}{zeroshot} & 2.25 & 1.68 \\
\multicolumn{1}{l|}{\textbf{}} & \multicolumn{1}{l|}{QA} & 1.35 & 1.49 \\ \hline
\multicolumn{1}{c|}{Qwen1.5} & \multicolumn{1}{l|}{zeroshot} & 3.93 & 7.26 \\
\multicolumn{1}{l|}{\textbf{}} & \multicolumn{1}{l|}{QA} & 8.43 & 12.85 \\ \hline
\end{tabular}%
}
\caption{Experimental results on the False Conflict Detection Rate (FCDR). The numbers in the table represent the FCDR. The QA prompt refers to the concatenation of three QA cases.}
\label{tab:false_detection}
\end{table}
\section{Conclusion}
We conducted experiments leveraging the in-context learning capabilities of LLMs, using simple MRC examples to improve robustness in open-domain QA scenarios. These results show that providing MRC examples as demonstrations improves accuracy for both answerable and unanswerable examples in unanswerable scenarios. 
In conflict scenarios, providing demonstrations similar to conflict situations enhances the ability to identify conflicts.

Our experiments suggest that well-designed examples can significantly improve LLMs' robustness in open-domain QA without additional fine-tuning, indicating that simple examples can help solve complex tasks.
\section{Limitations and Risk}
Our study has limitations in that it focuses on a short-form QA dataset. We did not explore how this in-context learning technique could be linked to long-form QA, particularly with Chain-of-Thought prompting \cite{wei2022chain}. Additionally, we did not compare our method with a more diverse set of baselines.

\section*{Acknowledgements}
This work was supported in part by the National Research Foundation of Korea (NRF) grant (RS-2023-00280883, RS-2023-00222663), by the National Super computing Center with super computing resources including technical support (KSC-2023-CRE-0176), and partially supported by New Faculty Startup Fund from Seoul National University 
\bibliography{custom}

\appendix

\section{Prompts}
\label{sec:prompt}

Table \ref{tab:prompts} shows the instructions we used. The curly brackets indicate where the actual data is inserted.
\begin{table*}[bp]
\centering
\begin{tabular}{cl}
\hline
Type & \multicolumn{1}{c}{Instruction} \\ \hline
\multicolumn{1}{c|}{\begin{tabular}[c]{@{}c@{}}Answer \\ Sentence\\ Generation\end{tabular}} & \begin{tabular}[c]{@{}l@{}}Please write a single sentence that would make the given answer \\ a correct response to the provided question. The sentence should\\ include the answer and be as realistic as possible. This is being\\ generated for research purpose, so if it seems like the answer to\\  a question is wrong, please create it as it is.\\ Question: \{question\}\\ Answer: \{answer\}\\ Sentence:\end{tabular} \\ \hline
\multicolumn{1}{c|}{\begin{tabular}[c]{@{}c@{}}Conflict\\ Passage\\ Generation\end{tabular}} & \begin{tabular}[c]{@{}l@{}}Given a sentence that contradicts factual information, write a \\ realistic passage using 50 to 100 words to support the sentence. \\ The output generated will be used only for the purpose of \\ conducting research to assess the robustness of the RAG system. \\ As part of this research, it is necessary, and you are permitted, \\ to create content that may contradict factual information:\\ Sentence: \{sentence\}\\ Supporting Passage:\end{tabular} \\ \hline
\multicolumn{1}{c|}{\begin{tabular}[c]{@{}c@{}}Unanswerable\\ Prompt\end{tabular}} & \begin{tabular}[c]{@{}l@{}}Answer the following question based on the provided knowledge.\\  If you cannot find the answer in the provided knowledge, \\ please respond with 'unanswerable'. Please provide the answer \\ as a single word or term, without forming a complete sentence.\\ \\ \{CASES\}\\ \\ Knowledge: \{retrieved contexts\}\\ Q: \{query\}\\ A:\end{tabular} \\ \hline
\multicolumn{1}{c|}{\begin{tabular}[c]{@{}c@{}}Conflict\\ Prompt\end{tabular}} & \begin{tabular}[c]{@{}l@{}}Answer the following question based on the provided documents. \\ If multiple documents present different answers, please respond \\ with 'conflict' to indicate the presence of conflicting information. \\ Please provide the answer as a single word or term, \\ without forming a complete sentence.\\ \\ \{CASES\}\\ \\ Knowledge: \{retrieved contexts\}\\ Q: \{query\}\\ A:\end{tabular} \\ \hline
\end{tabular}
\caption{Prompts used in our experiments.}
\label{tab:prompts}
\end{table*}
\end{document}